\documentclass[11pt,a4paper]{article}
\usepackage[hyperref]{acl2019}
\usepackage{times}
\usepackage{latexsym}
\usepackage{color}
\usepackage{multirow}
\usepackage{forest}
\usepackage{tikz}
\usepackage{algorithm}
\usepackage{algpseudocode}
\usepackage{amsmath}
\usepackage{graphics}
\usepackage{subfigure}
\usepackage{mwe}
\usepackage{amssymb}
\usepackage{bm}
\usepackage{amsmath}
\usepackage{makecell}
\usepackage{tabularx}
\usepackage{filecontents}
\usepackage{url}
\usepackage{pgfplots}
\usepackage{CJKutf8}
\pgfplotsset{width=10cm,compat=1.9}

\aclfinalcopy 

\definecolor{electricviolet}{rgb}{0.56, 0.0, 1.0}

\newcommand{\lexical}{\mathrm{lm}}
\newcommand{\form}{\mathrm{wf}}
\newcommand{\unrelated}{\mathrm{ur}}

\newcommand{\myshortcite}[1]{\citeauthor{#1} \shortcite{#1}}

\usetikzlibrary{arrows.meta,calc,decorations.markings,math,arrows.meta,patterns}
\usetikzlibrary{decorations.pathreplacing}
\usetikzlibrary{matrix}

\title{Shared-Private Bilingual Word Embeddings  \\ for Neural Machine Translation}

\author{Xuebo Liu$^\dagger$~~~Derek F. Wong$^\dagger$\thanks{~~Corresponding author}~~~Yang Liu$^\ddagger$~~~Lidia S. Chao$^\dagger$~~~Tong Xiao$^\mathsection$~~~Jingbo Zhu$^\mathsection$\\
 $^\dagger$NLP$^2$CT Lab / Department of Computer and Information Science, University of Macau, Macau \\
 $^\ddagger$Department of Computer Science and Technology, Tsinghua University, Beijing, China\\
$^\mathsection$Northeastern University, Shenyang, China \\
 {\tt nlp2ct.xuebo@gmail.com, \{derekfw,lidiasc\}@um.edu.mo,} \\
 {\tt liuyang2011@tsinghua.edu.cn,} 
 {\tt \{xiaotong,zhujingbo\}@mail.neu.edu.cn}\\}
 
\date{}

\begin{document}
\maketitle
\begin{abstract}
Word embedding is central to neural machine translation (NMT), which has attracted intensive research interest in recent years. 
In NMT, the source embedding plays the role of the entrance while the target embedding acts as the terminal. 
These layers occupy most of the model parameters for representation learning.
Furthermore, they indirectly interface via a soft-attention mechanism, which makes them comparatively isolated. 
In this paper, we propose \emph{shared-private} bilingual word embeddings, which give a closer relationship between the source and target embeddings, and which also reduce the number of model parameters.
For similar source and target words, their embeddings tend to share a part of the features and they cooperatively learn these common representation units.
Experiments on 5 language pairs belonging to 6 different language families and written in 5 different alphabets demonstrate that the proposed model provides a significant performance boost over the strong baselines with dramatically fewer model parameters.
\end{abstract}

\begin{CJK}{UTF8}{gbsn}
\section{Introduction}
With the introduction of ever more powerful architectures, neural machine translation (NMT) has become the most promising machine translation method \cite{kalchbrenner2013recurrent,sutskever2014sequence,bahdanau2014neural}.
For word representation, different architectures---including, but not limited to, recurrence-based \cite{Chen:2018vf}, convolution-based \cite{DBLP:journals/corr/GehringAGYD17} and transformation-based \cite{DBLP:journals/corr/VaswaniSPUJGKP17} NMT models---have been taking advantage of the distributed word embeddings to capture the syntactic and semantic properties of words \cite{Turian:2010vi}.

NMT usually utilizes three matrices to represent source embeddings, target input embeddings, and target output embeddings (also known as pre-softmax weight), respectively. 
These embeddings occupy most of the model parameters, which constrains the improvements of NMT because the recent methods become increasingly memory-hungry \cite{DBLP:journals/corr/VaswaniSPUJGKP17,Chen:2018vf}.\footnote{For the purpose of smoothing gradients, a very large batch size is needed during training.} 
Even though converting words into sub-word units \cite{DBLP:journals/corr/SennrichHB15}, nearly 55\% of model parameters are used for word representation in the Transformer model \cite{DBLP:journals/corr/VaswaniSPUJGKP17}.

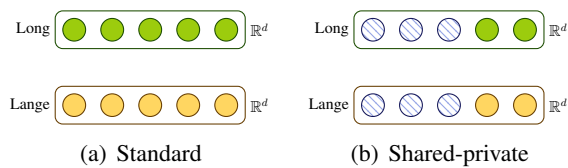
\begin{figure}[t] 
    \centering
    \subfigure[Standard]{
    \scalebox{0.5}{\begin{tikzpicture}

    \definecolor{Eycolor}{rgb}{.59,.65,.90}
	\definecolor{A1color}{rgb}{.61,.80,.05}
	\definecolor{A2color}{rgb}{1.0,.84,.38}
	
	\definecolor{Eyxcolor}{rgb}{.08,.09,.35}
	\definecolor{A1xcolor}{rgb}{.11,.30,.0}
	\definecolor{A2xcolor}{rgb}{.40,.24,.0}
	
    \coordinate (A1start) at (0,2);
    \coordinate (A1end) at (5,3);
    \node [left] at (0,2.5) {\large{Long}};
    \node [right] at (5,2.5) {\large{$\mathbb{R}^d$}};
    \draw [A1xcolor, fill=white, thick, rounded corners=2mm] (A1start) rectangle (A1end);
    \foreach \x in {0,...,4} 
    	{\draw [A1xcolor, fill=A1color, thick] (\x+0.5,2.5) circle [radius=0.3];}
    	
    \coordinate (A2start) at (0,0);
    \coordinate (A2end) at (5,1);
    \node [left] at (0,0.5) {\large{Lange}};
    \node [right] at (5,0.5) {\large{$\mathbb{R}^d$}};
    \draw [A2xcolor, fill=white, thick, rounded corners=2mm] (A2start) rectangle (A2end);
    \foreach \x in {0,...,4} 
    	{\draw [A2xcolor, fill=A2color, thick] (\x+0.5,0.5) circle [radius=0.3];}

\end{tikzpicture}}
    \label{fig:standard}}
    \hfill
    \subfigure[Shared-private]{
    \scalebox{0.5}{\begin{tikzpicture}

    \definecolor{Eycolor}{rgb}{.59,.65,.90}
	\definecolor{A1color}{rgb}{.61,.80,.05}
	\definecolor{A2color}{rgb}{1.0,.84,.38}
	
	\definecolor{Eyxcolor}{rgb}{.08,.09,.35}
	\definecolor{A1xcolor}{rgb}{.11,.30,.0}
	\definecolor{A2xcolor}{rgb}{.40,.24,.0}
	
    \coordinate (A1start) at (0,2);
    \coordinate (A1end) at (5,3);
    \node [left] at (0,2.5) {\large{Long}};
    \node [right] at (5,2.5) {\large{$\mathbb{R}^d$}};
    \draw [A1xcolor, fill=white, thick, rounded corners=2mm] (A1start) rectangle (A1end);
    \foreach \x in {3,...,4} 
    	{\draw [A1xcolor, fill=A1color, thick] (\x+0.5,2.5) circle [radius=0.3];}
    \foreach \x in {0,...,2} 
    	{\draw [Eyxcolor, pattern=north west lines,pattern color=Eycolor, thick] (\x+0.5,2.5) circle [radius=0.3];}
    	
    \coordinate (A2start) at (0,0);
    \coordinate (A2end) at (5,1);
    \node [left] at (0,0.5) {\large{Lange}};
    \node [right] at (5,0.5) {\large{$\mathbb{R}^d$}};
    \draw [A2xcolor, fill=white, thick, rounded corners=2mm] (A2start) rectangle (A2end);
    \foreach \x in {3,...,4} 
    	{\draw [A2xcolor, fill=A2color, thick] (\x+0.5,0.5) circle [radius=0.3];}
    \foreach \x in {0,...,2} 
    	{\draw [Eyxcolor, pattern=north west lines,pattern color=Eycolor, thick] (\x+0.5,0.5) circle [radius=0.3];}
    	
%
%

\end{tikzpicture}}
    \label{fig:lexical1}}
    \caption{Comparison between (a) standard word embeddings and (b) shared-private word embeddings. In (a), the English word ``Long'' and the German word ``Lange'', which have similar lexical meanings, are represented by two private $d$-dimension vectors. While in (b), the two word embeddings are made up of two parts, indicating the shared (lined nodes) and the private (unlined nodes) features. This enables the two words to make use of common representation units, leading to a closer relationship between them.}
    \label{fig:shareEmb}
\end{figure}

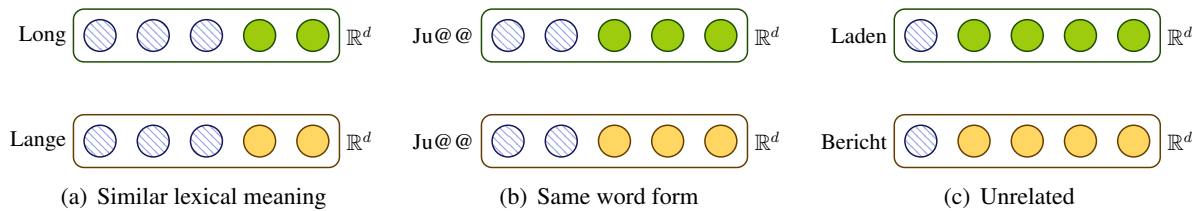
\begin{figure*}[t] 
    \centering
    \subfigure[Similar lexical meaning]{
    \scalebox{0.7}{\begin{tikzpicture}

    \definecolor{Eycolor}{rgb}{.59,.65,.90}
	\definecolor{A1color}{rgb}{.61,.80,.05}
	\definecolor{A2color}{rgb}{1.0,.84,.38}
	
	\definecolor{Eyxcolor}{rgb}{.08,.09,.35}
	\definecolor{A1xcolor}{rgb}{.11,.30,.0}
	\definecolor{A2xcolor}{rgb}{.40,.24,.0}
	
    \coordinate (A1start) at (0,2);
    \coordinate (A1end) at (5,3);
    \node [left] at (0,2.5) {\large{Long}};
    \node [right] at (5,2.5) {\large{$\mathbb{R}^d$}};
    \draw [A1xcolor, fill=white, thick, rounded corners=2mm] (A1start) rectangle (A1end);
    \foreach \x in {3,...,4} 
    	{\draw [A1xcolor, fill=A1color, thick] (\x+0.5,2.5) circle [radius=0.3];}
    \foreach \x in {0,...,2} 
    	{\draw [Eyxcolor, pattern=north west lines,pattern color=Eycolor, thick] (\x+0.5,2.5) circle [radius=0.3];}
    	
    \coordinate (A2start) at (0,0);
    \coordinate (A2end) at (5,1);
    \node [left] at (0,0.5) {\large{Lange}};
    \node [right] at (5,0.5) {\large{$\mathbb{R}^d$}};
    \draw [A2xcolor, fill=white, thick, rounded corners=2mm] (A2start) rectangle (A2end);
    \foreach \x in {3,...,4} 
    	{\draw [A2xcolor, fill=A2color, thick] (\x+0.5,0.5) circle [radius=0.3];}
    \foreach \x in {0,...,2} 
    	{\draw [Eyxcolor, pattern=north west lines,pattern color=Eycolor, thick] (\x+0.5,0.5) circle [radius=0.3];}
    	
%
%

\end{tikzpicture}}
    \label{fig:lexical2}}
    \hfill
    \subfigure[Same word form]{
    \scalebox{0.7}{\begin{tikzpicture}

    \definecolor{Eycolor}{rgb}{.59,.65,.90}
	\definecolor{A1color}{rgb}{.61,.80,.05}
	\definecolor{A2color}{rgb}{1.0,.84,.38}
	
	\definecolor{Eyxcolor}{rgb}{.08,.09,.35}
	\definecolor{A1xcolor}{rgb}{.11,.30,.0}
	\definecolor{A2xcolor}{rgb}{.40,.24,.0}
	
    \coordinate (A1start) at (0,2);
    \coordinate (A1end) at (5,3);
    \node [left] at (0,2.5) {\large{Ju@@}};
    \node [right] at (5,2.5) {\large{$\mathbb{R}^d$}};
    \draw [A1xcolor, fill=white, thick, rounded corners=2mm] (A1start) rectangle (A1end);
    \foreach \x in {2,...,4} 
    	{\draw [A1xcolor, fill=A1color, thick] (\x+0.5,2.5) circle [radius=0.3];}
    \foreach \x in {0,...,1} 
    	{\draw [Eyxcolor, pattern=north west lines,pattern color=Eycolor, thick] (\x+0.5,2.5) circle [radius=0.3];}
    	
    \coordinate (A2start) at (0,0);
    \coordinate (A2end) at (5,1);
    \node [left] at (0,0.5) {\large{Ju@@}};
    \node [right] at (5,0.5) {\large{$\mathbb{R}^d$}};
    \draw [A2xcolor, fill=white, thick, rounded corners=2mm] (A2start) rectangle (A2end);
    \foreach \x in {2,...,4} 
    	{\draw [A2xcolor, fill=A2color, thick] (\x+0.5,0.5) circle [radius=0.3];}
    \foreach \x in {0,...,1} 
    	{\draw [Eyxcolor, pattern=north west lines,pattern color=Eycolor, thick] (\x+0.5,0.5) circle [radius=0.3];}
    	
%
%

\end{tikzpicture}}
    \label{fig:form}}
    \hfill
    \subfigure[Unrelated]{
    \scalebox{0.7}{\begin{tikzpicture}

    \definecolor{Eycolor}{rgb}{.59,.65,.90}
	\definecolor{A1color}{rgb}{.61,.80,.05}
	\definecolor{A2color}{rgb}{1.0,.84,.38}
	
	\definecolor{Eyxcolor}{rgb}{.08,.09,.35}
	\definecolor{A1xcolor}{rgb}{.11,.30,.0}
	\definecolor{A2xcolor}{rgb}{.40,.24,.0}
	
    \coordinate (A1start) at (0,2);
    \coordinate (A1end) at (5,3);
    \node [left] at (0,2.5) {\large{Laden}};
    \node [right] at (5,2.5) {\large{$\mathbb{R}^d$}};
    \draw [A1xcolor, fill=white, thick, rounded corners=2mm] (A1start) rectangle (A1end);
    \foreach \x in {1,...,4} 
    	{\draw [A1xcolor, fill=A1color, thick] (\x+0.5,2.5) circle [radius=0.3];}
    \foreach \x in {0,...,0} 
    	{\draw [Eyxcolor, pattern=north west lines,pattern color=Eycolor,thick] (\x+0.5,2.5) circle [radius=0.3];}
    	
    \coordinate (A2start) at (0,0);
    \coordinate (A2end) at (5,1);
    \node [left] at (0,0.5) {\large{Bericht}};
    \node [right] at (5,0.5) {\large{$\mathbb{R}^d$}};
    \draw [A2xcolor, fill=white, thick, rounded corners=2mm] (A2start) rectangle (A2end);
    \foreach \x in {1,...,4} 
    	{\draw [A2xcolor, fill=A2color, thick] (\x+0.5,0.5) circle [radius=0.3];}
    \foreach \x in {0,...,0} 
    	{\draw [Eyxcolor, pattern=north west lines,pattern color=Eycolor, thick] (\x+0.5,0.5) circle [radius=0.3];}
    	
%
%

\end{tikzpicture}}
    \label{fig:unrelated}}
    \hfill
    \caption{Shared-private bilingual word embeddings perform between the source and target words or sub-words (a) {with similar lexical meaning}, (b) with same word form, and (c) without any relationship. Different sharing mechanisms are adapted into different relationship categories. This strikes the right balance between capturing monolingual and bilingual characteristics. The closeness of relationship decides the portion of features to be used for sharing. Words with similar lexical meaning tend to share more features, followed by the words with the same word form, and then the unrelated words, as illustrated by the lined nodes.}
    \label{fig:shareEmbAll}
\end{figure*}


To overcome this difficulty, several methods are proposed to reduce the parameters used for word representation of NMT. 
\myshortcite{Press:2017ug} propose two weight tying (WT) methods, called decoder WT and three-way WT, to substantially reduce the parameters of the word embeddings.
Decoder WT ties the target input embedding and target output embedding, which has become the new \emph{de facto} standard of practical NMT \cite{Sennrich:2017uo}. 
Three-way WT uses only one matrix to represent the three word embeddings, where the source and target words that have the same word form tend to share a word vector. 
This method can also be adapted to sub-word NMT with a shared source-target sub-word vocabulary and it performs well in language pairs with many of the same characters, such as English-German and English-French \cite{DBLP:journals/corr/VaswaniSPUJGKP17}.
Unfortunately, this method is not applicable to languages that are written in different alphabets, such as Chinese-English \cite{Hassan:2018vv}. 

Another challenge facing the source and target word embeddings of NMT is the lack of interactions.
This degrades the attention performance, leading to some unaligned translations that hurt the translation quality.
Hence, \myshortcite{Kuang:2017vk} propose to bridge the source and target embeddings, which brings better attention to the related source and target words.
Their method is applicable to any language pairs, providing a tight interaction between the source and target word pairs.
However, their method requires additional components and model parameters.



In this work, we aim to enhance the word representations and the interactions between the source and target words, while using even fewer parameters. 
To this end, we present a language-independent method, which is called shared-private bilingual word embeddings, to share a part of the embeddings of a pair of source and target words that have some common characteristics (i.e. similar words should have similar vectors).
Figure~\ref{fig:shareEmb} illustrates the difference between the standard word embeddings and shared-private word embeddings of NMT.
In the proposed method, each source (or target) word is represented by a word embedding that consists of the shared features and the private features.
The shared features can also be regarded as the prior alignments connecting the source and target words.
The private features allow the words to better learn the monolingual characteristics.
Meanwhile, the features shared by the source and target embeddings result in a significant reduction of the number of parameters used for word representations.
The experimental results on 6 translation datasets of different scales show that our model with fewer parameters yields consistent improvements over the strong Transformer baselines.

\section{Approach}
In monolingual vector space, similar words tend to have commonalities in the same dimensions of their word vectors \cite{mikolov2013efficient}. 
These commonalities include: (1) a similar degree (value) of the same dimension and (2) a similar positive or negative correlation of the same dimension.
Many previous works have noticed this phenomenon and have proposed to use shared vectors to represent similar words in monolingual vector space toward model compression \cite{DBLP:journals/corr/LiQYL16,Zhang:2017wt,Li:2017td}.

Motivated by these works, in NMT, we assume that the source and target words that have similar characteristics should also have similar vectors.
Hence, we propose to perform this sharing technique in bilingual vector space.
More precisely, we share the features (dimensions) between the paired source and target embeddings (vectors).
However, in contrast to the previous studies, we also model the private features of the word embedding to preserve the private characteristics of words for source and target languages. 
The private features allow the words to better learn the monolingual characteristics.
Meanwhile, we also propose to adopt different sharing mechanisms among the word pairs, which will be described in the following sections.

In the Transformer architecture, the shared features between the source and target embeddings always contribute to the calculation of the attention weight.\footnote{Based on the dot-product attention mechanism, the attention weight between the source and target embeddings is the sum of the dot-product of their features.}
This results in paying more attention strength on the pair of related words.
With the help of residual connections, the high-level representations can also benefit from the shared features of the topmost embedding layers.
Both qualitative and quantitative analyses show the effectiveness on the translation tasks.

\subsection{Shared-Private Bilingual Word Embeddings}
Standard NMT jointly learns to translate and align, which has achieved remarkable results \cite{bahdanau2014neural}.
In NMT, the intention is to identify the translation relationships between the source and target words.
To simplify the model, we propose to divide the relationships into three main categories between a pair of source and target words: (1) words with similar lexical meaning (abbreviated as $\lexical$), (2) words with same word form (abbreviated as $\form$), and (3) unrelated words (abbreviated as $\unrelated$).
Figure~\ref{fig:shareEmbAll} shows some examples of these different relationship categories.
The number of the shared features of the word embeddings is decided by their relationships.

Before presenting the pairing process in detail, we first introduce the constraints to the proposed method for convenience:
\begin{itemize}
    \item Each source word is only allowed to share the features with a single target word, and vice versa.\footnote{We investigate the effect of synonym in the experiment section.}
    \item Each source word preferentially shares features with the target word that has similar lexical meaning, followed by the word with same word form, and then unrelated words.
\end{itemize}

\subsubsection{Words with Similar Lexical Meaning}
As shown in Figure~\ref{fig:lexical2}, the English word ``Long'' and the German word ``Lange'', which have similar meaning, tend to share more common features of their embeddings.
In our model, the source and target words with alignment links are regarded as parallel words that are the translation of each other.
According to the word frequency, each source word $x$ is paired with a target aligned word $\hat{y}$ that has the highest alignment probability among the candidates, and is computed as follows:
\begin{eqnarray}
\hat{y} = \mathop{\arg\max}_{y \in a(x)}\mathrm{log}A(y|x)
\label{equ:align}
\end{eqnarray}
where $a(\cdot)$ denotes the set of aligned candidates. 
It is worth noting the target words that have been paired with the source words cannot be used as candidates.
$A(\cdot|\cdot)$ denotes the alignment probability. 
These can be obtained by either the intrinsic attention mechanism \cite{bahdanau2014neural} or unsupervised word aligner \cite{Dyer:2013wv}.
\subsubsection{Words with Same Word Form}
As shown in Figure~\ref{fig:form}, the sub-word ``Ju@@'' simultaneously exists in English and German sentences.
This kind of word tends to share a medium number of features of the word embeddings.
Most of the time, the source and target words with the same word form also share similar lexical meaning.
This category of words generally includes Arabic numbers, punctuations, 
named entities, cognates and loanwords.
However, there are some bilingual homographs where the words in the source and target languages look the same but have completely different meanings. 
For example, the German word ``Gift'' means ``Poison'' in English.
That is the reason we propose to first pair the words with similar lexical meaning instead of those words with same word forms.
This might be the potential limitation of the three-way WT method \cite{Press:2017ug}, where  words with the same word form indiscriminately share the same word embedding.

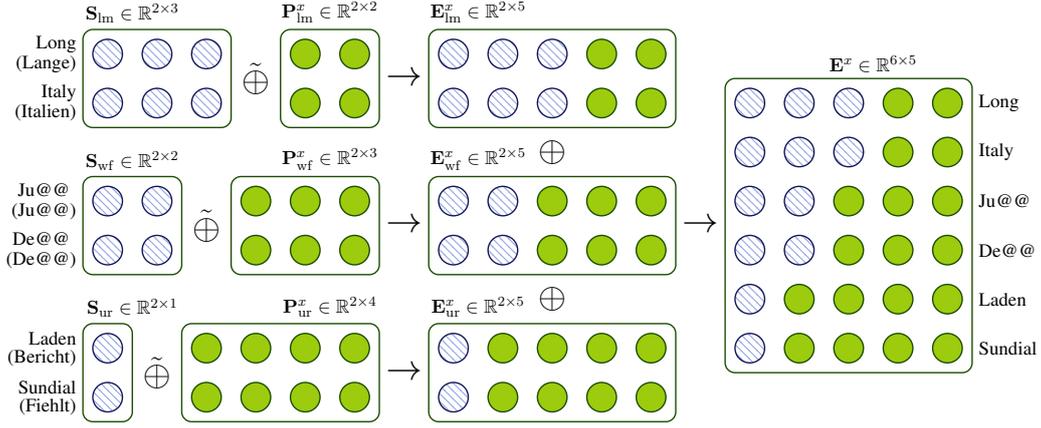
\begin{figure*}[t] 
    \centering
    \scalebox{0.65}{\begin{tikzpicture}

    \definecolor{Eycolor}{rgb}{.59,.65,.90}
	\definecolor{A1color}{rgb}{.61,.80,.05}
	\definecolor{A2color}{rgb}{1.0,.84,.38}
	
	\definecolor{Eyxcolor}{rgb}{.08,.09,.35}
	\definecolor{A1xcolor}{rgb}{.11,.30,.0}
	\definecolor{A2xcolor}{rgb}{.40,.24,.0}
	
	\coordinate (A1start) at (13,3-1);
    \coordinate (A1end) at (18,9-1);
    \draw [A1xcolor, fill=white, thick, rounded corners=2mm] (A1start) rectangle (A1end);
    \node [above] at (16,9-1) {$\mathbf{E}^x \in \mathbb{R}^{6 \times 5}$};
	
    \coordinate (A1start) at (0,1+8-2);
    \coordinate (A1end) at (3,3+8-2);
    \coordinate (A2start) at (4,1+8-2);
    \coordinate (A2end) at (6,3+8-2);
    \coordinate (A3start) at (7,1+8-2);
    \coordinate (A3end) at (12,3+8-2);    
    
    \node [left] at (0,2.7+8-2) {Long};
    \node [right] at (18,2.5+6-1) {Long};
    \node [left] at (0,2.3+8-2) {(Lange)};
    \node [left] at (0,1.7+8-2) {Italy};
    \node [right] at (18,1.5+6-1) {Italy};
    \node [left] at (0,1.3+8-2) {(Italien)};

    \node [above] at (8,3+8-2) {$\mathbf{E}_{\lexical}^x \in \mathbb{R}^{2 \times 5}$};
    \node [above] at (1,3+8-2) {$\mathbf{S}_{\lexical} \in \mathbb{R}^{2 \times 3}$};
    \node [above] at (5,3+8-2) {$\mathbf{P}_{\lexical}^x \in \mathbb{R}^{2 \times 2}$};
    \node at (3.5,2+8-2) {\huge{$\tilde{\oplus}$}};
    \node at (6.5,2+8-2) {\huge{$\rightarrow$}};
    \node at (9.5,4-1+0.5) {\huge{$\oplus$}};

    \draw [A1xcolor, fill=white, thick, rounded corners=2mm] (A1start) rectangle (A1end);
    \draw [A1xcolor, fill=white, thick, rounded corners=2mm] (A2start) rectangle (A2end);
    \draw [A1xcolor, fill=white, thick, rounded corners=2mm] (A3start) rectangle (A3end);
    \foreach \x in {4,...,5,10,11} 
    	{\draw [A1xcolor, fill=A1color, thick] (\x+0.5,2.5+8-2) circle [radius=0.3];
    	 \draw [A1xcolor, fill=A1color, thick] (\x+0.5,1.5+8-2) circle [radius=0.3];
    	}
        \foreach \x in {16,17} 
    	{\draw [A1xcolor, fill=A1color, thick] (\x+0.5,4.5+4-1) circle [radius=0.3];
    	 \draw [A1xcolor, fill=A1color, thick] (\x+0.5,3.5+4-1) circle [radius=0.3];
    	}
    	
    \foreach \x in {0,...,2} 
    	{\draw [Eyxcolor, pattern=north west lines,pattern color=Eycolor, thick] (\x+0.5,2.5+8-2) circle [radius=0.3];
    	\draw [Eyxcolor, pattern=north west lines,pattern color=Eycolor, thick] (\x+0.5,1.5+8-2) circle [radius=0.3];
    	}
        \foreach \x in {7,...,9} 
    	{\draw [Eyxcolor, pattern=north west lines,pattern color=Eycolor, thick] (\x+0.5,2.5+8-2) circle [radius=0.3];
    	\draw [Eyxcolor, pattern=north west lines,pattern color=Eycolor, thick] (\x+0.5,1.5+8-2) circle [radius=0.3];
    	}
    	 \foreach \x in {13,...,15} 
    	{\draw [Eyxcolor, pattern=north west lines,pattern color=Eycolor, thick] (\x+0.5,4.5+4-1) circle [radius=0.3];
    	\draw [Eyxcolor, pattern=north west lines,pattern color=Eycolor, thick] (\x+0.5,3.5+4-1) circle [radius=0.3];
    	}

    \coordinate (A1start) at (0,1+4-1);
    \coordinate (A1end) at (2,3+4-1);
    \coordinate (A2start) at (3,1+4-1);
    \coordinate (A2end) at (6,3+4-1);
    \coordinate (A3start) at (7,1+4-1);
    \coordinate (A3end) at (12,3+4-1);    
    
    \node [left] at (0,2.7+4-1) {Ju@@\textcolor{white}{(}};
    \node [left] at (0,1.7+4-1) {De@@\textcolor{white}{(}};
    \node [left] at (0,2.3+4-1) {(Ju@@)};
    \node [right] at (18,2.5+4-1) {Ju@@};
    \node [right] at (18,1.5+4-1) {De@@};
    \node [left] at (0,1.3+4-1) {(De@@)};
    \node [above] at (8,3+4-1) {$\mathbf{E}_{\form}^x \in \mathbb{R}^{2 \times 5}$};
    \node [above] at (1,3+4-1) {$\mathbf{S}_{\form} \in \mathbb{R}^{2 \times 2}$};
    \node [above] at (5,3+4-1) {$\mathbf{P}_{\form}^x \in \mathbb{R}^{2 \times 3}$};
    \node at (2.5,2+4-1) {\huge{$\tilde{\oplus}$}};
    \node at (6.5,2+4-1) {\huge{$\rightarrow$}};
    \node at (12.5,2+4-1) {\huge{$\rightarrow$}};
    \node at (9.5,4+4-1-0.5) {\huge{$\oplus$}};

    \draw [A1xcolor, fill=white, thick, rounded corners=2mm] (A1start) rectangle (A1end);
    \draw [A1xcolor, fill=white, thick, rounded corners=2mm] (A2start) rectangle (A2end);
    \draw [A1xcolor, fill=white, thick, rounded corners=2mm] (A3start) rectangle (A3end);
    \foreach \x in {3,...,5,9,10,11,15,16,17} 
    	{\draw [A1xcolor, fill=A1color, thick] (\x+0.5,2.5+4-1) circle [radius=0.3];
    	 \draw [A1xcolor, fill=A1color, thick] (\x+0.5,1.5+4-1) circle [radius=0.3];
    	}
    	
    \foreach \x in {0,...,1} 
    	{\draw [Eyxcolor, pattern=north west lines,pattern color=Eycolor, thick] (\x+0.5,2.5+4-1) circle [radius=0.3];
    	\draw [Eyxcolor, pattern=north west lines,pattern color=Eycolor, thick] (\x+0.5,1.5+4-1) circle [radius=0.3];
    	}
        \foreach \x in {7,...,8,13,14} 
    	{\draw [Eyxcolor, pattern=north west lines,pattern color=Eycolor, thick] (\x+0.5,2.5+4-1) circle [radius=0.3];
    	\draw [Eyxcolor, pattern=north west lines,pattern color=Eycolor, thick] (\x+0.5,1.5+4-1) circle [radius=0.3];
    	}

    \coordinate (A1start) at (0,1);
    \coordinate (A1end) at (1,3);
    \coordinate (A2start) at (2,1);
    \coordinate (A2end) at (6,3);
    \coordinate (A3start) at (7,1);
    \coordinate (A3end) at (12,3);    
    
    \node [left] at (0,2.7) {Laden};
    \node [left] at (0,2.3) {(Bericht)};
    \node [left] at (0,1.7) {Sundial};
    \node [left] at (0,1.3) {(Fiehlt)};
    \node [right] at (18,2.5+2-1) {Laden};
    \node [right] at (18,1.5+2-1) {Sundial};
    \node [above] at (8,3) {$\mathbf{E}_{\unrelated}^x \in \mathbb{R}^{2 \times 5}$};
    \node [above] at (1,3) {$\mathbf{S}_{\unrelated} \in \mathbb{R}^{2 \times 1}$};
    \node [above] at (5,3) {$\mathbf{P}_{\unrelated}^x \in \mathbb{R}^{2 \times 4}$};
    \node at (1.5,2) {\huge{$\tilde{\oplus}$}};
    \node at (6.5,2) {\huge{$\rightarrow$}};

    \draw [A1xcolor, fill=white, thick, rounded corners=2mm] (A1start) rectangle (A1end);
    \draw [A1xcolor, fill=white, thick, rounded corners=2mm] (A2start) rectangle (A2end);
    \draw [A1xcolor, fill=white, thick, rounded corners=2mm] (A3start) rectangle (A3end);
    \foreach \x in {2,...,5,8,9,10,11} 
    	{\draw [A1xcolor, fill=A1color, thick] (\x+0.5,2.5) circle [radius=0.3];
    	 \draw [A1xcolor, fill=A1color, thick] (\x+0.5,1.5) circle [radius=0.3];
    	}
    	    \foreach \x in {14,...,17} 
    	{\draw [A1xcolor, fill=A1color, thick] (\x+0.5,2.5+2-1) circle [radius=0.3];
    	 \draw [A1xcolor, fill=A1color, thick] (\x+0.5,1.5+2-1) circle [radius=0.3];
    	}
    	
    \foreach \x in {0,...,0} 
    	{\draw [Eyxcolor, pattern=north west lines,pattern color=Eycolor, thick] (\x+0.5,2.5) circle [radius=0.3];
    	\draw [Eyxcolor, pattern=north west lines,pattern color=Eycolor, thick] (\x+0.5,1.5) circle [radius=0.3];
    	}
        \foreach \x in {7,...,7} 
    	{\draw [Eyxcolor, pattern=north west lines,pattern color=Eycolor, thick] (\x+0.5,2.5) circle [radius=0.3];
    	\draw [Eyxcolor, pattern=north west lines,pattern color=Eycolor, thick] (\x+0.5,1.5) circle [radius=0.3];
    	}
    	        \foreach \x in {13,...,13} 
    	{\draw [Eyxcolor, pattern=north west lines,pattern color=Eycolor, thick] (\x+0.5,2.5+2-1) circle [radius=0.3];
    	\draw [Eyxcolor, pattern=north west lines,pattern color=Eycolor, thick] (\x+0.5,1.5+2-1) circle [radius=0.3];
    	}

\end{tikzpicture}}
    \caption{The example of assembling the source word embedding matrix. The words in parentheses denote the paired words sharing features with them.}
    \label{fig:preemb}
\end{figure*}

\subsubsection{Unrelated Words}
We regard source and target words that cannot be paired with each other as unrelated words.
Figure~\ref{fig:unrelated} shows an example of a pair of unrelated words.
This category is mainly composed of low-frequency words, such as misspelled words, special characters, and foreign words.
In standard NMT, the embeddings of low-frequency words are usually inadequately trained, resulting in a poor word representation.
These words are often treated as noises and they are generally ignored by the NMT systems \cite{Feng:vt}.
Motivated by the frequency clustering methods proposed by \citet{chen2015strategies} where they cluster the words with similar frequency for training a hierarchical language model, in this work, we propose to use a small vector to model the possible features that might be shared between the source and target words which are unrelated but having similar word frequencies.
In addition, it can be regarded as a way to improve the robustness of learning the embeddings of low-frequency words because of the noisy dimensions \cite{Wang:2018vb}.
\begin{table*}[ht]
\small
    \centering
    \scalebox{1.0}{
    \begin{tabularx}{\textwidth}{l|lrrr|l|lllll}
        \hline
        Architecture     &\textcolor{white}{ }Zh$\Rightarrow$En  &Params &Emb.& Red. &Dev. & MT02 & MT03 & MT04  & MT08 &All\\ \hline \hline 
        SMT*      &- &-  &- &- &34.00     &35.81     &34.70    &37.15   &25.28 &33.39  \\ \hline
        \multirow{4}{*}{RNNsearch*}  &Vanilla &74.8M &55.8M &0\% &35.92 &37.88 &36.21 &38.83  &26.30 &34.81 \\
        &Source bridging &78.5M &55.8M &0\% &36.79 &38.71 &37.24 &40.28  &27.40 &35.91 \\
        &Target bridging &76.6M &55.8M &0\% &36.69 &39.04 &37.63 &40.41  &27.98 &36.27 \\ 
        &Direct bridging &78.9M &55.8M &0\% &36.97 &39.77 &38.02 &40.83  &27.85 &36.62 \\ \hline
        \multirow{4}{*}{Transformer} &Vanilla &90.2M  &46.1M &0\%  &41.37 &42.53 &40.25 &43.58 &32.89 &40.33 \\ 
        &Direct bridging &90.5M  &46.1M &0\% &41.67 &42.89 &41.34 &43.56  &32.69 &40.54 \\ 
        &Decoder WT &74.9M  &30.7M &33.4\% &41.90 &43.02 &41.89 &43.87  &32.62 &40.82 \\ 
        &\emph{Shared-private} &62.8M  &18.7M &59.4\% &42.57$^\uparrow$ &43.73$^\uparrow$ &41.99$^\uparrow$ &44.53$^\uparrow$  &33.81$^\Uparrow$ &41.61$^\Uparrow$ \\  \hline
    \end{tabularx}}
    \caption{Results on the NIST Chinese-English translation task. ``Params'' denotes the number of model parameters. ``Emb.'' represents the number of parameters used for word representation. ``Red.'' represents the reduction rate of the standard size. The results of SMT* and RNNsearch* are reported by \myshortcite{Kuang:2017vk}  with the same datasets and vocabulary settings. ``$\uparrow$'' indicates the result is significantly better than that of the vanilla Transformer ($p < 0.01$), while ``$\Uparrow$'' indicates the result is significantly better than that of all other Transformer models ($p < 0.01$). All significance tests are measured by paired bootstrap resampling~\cite{koehn2004statistical}.} 
    \label{tab:zhenresults}    
\end{table*}

\begin{table}[ht]
\small
    \centering
    \scalebox{1.0}{
    \begin{tabularx}{0.47\textwidth}{@{\extracolsep{\fill}}l|l|lr|l}
        \hline
        En$\Rightarrow$De &Params &Emb.& Red. &BLEU\\ \hline \hline 
        Vanilla &98.7M  &54.5M &0\% &27.62  \\ 
        Direct bridging &98.9M  &54.5M &0\% &27.79  \\ 
        Decoder WT &80.4M  &36.2M &33.6\% &27.51  \\ 
        Three-way WT &63.1M  &18.9M &65.3\% &27.39  \\ \hline \hline
        \emph{Shared-private} &65.0M  &20.9M &63.1\% &28.06$^\ddagger$ \\ \hline
    \end{tabularx}}
    \caption{Results on the WMT English-German translation task. ``$\ddagger$'' indicates the result is significantly better than the vanilla Transformer model ($p < 0.05$).} 
    \label{tab:enderesults}    
\end{table}

\subsection{Implementation}
Before looking up embedding at each training step, the source and target embedding matrix are assembled by the sub-embedding matrices.
As shown in Figure~\ref{fig:preemb}, the source embedding $\mathbf{E}^x \in {\mathbb{R}^{|V| \times d}}$  is computed as follows::
\begin{eqnarray}
\mathbf{E}^x = \mathbf{E}^x_{\mathrm{\lexical}} \oplus \mathbf{E}^x_{\mathrm{\form}} \oplus \mathbf{E}^x_{\mathrm{\unrelated}} 
\end{eqnarray}
where $\oplus$ is the row concatenation operator. $\mathbf{E}^x_{(\cdot)} \in {\mathbb{R}^{|V_{(\cdot)}| \times d}}$ represents the word embeddings of the source words belong to different categories, e.g. $\lexical$ represents the words with similar lexical meaning.
$|V_{(\cdot)}|$ denotes the vocabulary size of the corresponding category.

The process of feature sharing is also implemented by matrix concatenation. For example, the embedding matrices of the source words with similar lexical meaning are computed as follows:
\begin{eqnarray}
\mathbf{E}^x_{\lexical} = \mathbf{S}_{\lexical} \tilde{\oplus} \mathbf{P}^x_{\lexical}
\end{eqnarray}
where $\tilde{\oplus}$ is the column concatenation operator. 
$\mathbf{S}_{\lexical} \in {\mathbb{R}^ {  |V_{\lexical}| \times \lambda_{\lexical}d}}$ represent the word embeddings of the shared features, where $\lambda_{\lexical}$ denotes the proportion of the features for sharing in this relationship category.
$\mathbf{P}^{x}_{\lexical} \in {\mathbb{R}^{|V_{\lexical}| \times (1-\lambda_{\lexical})d  }}$ represent the word embeddings of the private features.

Similar to the target word embedding.
These matrix concatenation operations, which have low computational complexity, are very cheap to the whole NMT computation process.
We also empirically find both the training speed and decoding speed are not influenced with the introduction of the proposed method.

\end{CJK}

\section{Experiments}
We carry out our experiments on the small-scale IWSLT'17 \{Arabic (Ar), Japanese (Ja), Korean (Ko), Chinese (Zh)\}-to-English (En) translation tasks, medium-scale NIST Chinese-English (Zh-En) translation task, and large-scale WMT'14 English-German (En-De) translation task. 

For the IWSLT \{Ar, Ja, Ko, Zh\}-to-En translation tasks, there are respectively 236K, 234K, 227K, and 235K sentence pairs in each training set.\footnote{\url{https://wit3.fbk.eu/mt.php?release=2017-01-trnted}}
The validation set is IWSLT17.TED.tst2014 and the test set is IWSLT17.TED.tst2015.
For each language, we learn a BPE model with 16K merge operations \cite{DBLP:journals/corr/SennrichHB15}.

For the NIST Zh-En translation task, the training corpus consists of 1.25M sentence pairs with 27.9M Chinese words and 34.5M English words.
We use the NIST MT06 dataset as the validation set and the test sets are the NIST MT02, MT03, MT04, MT05, MT08 datasets.
To compare with the recent works, the vocabulary size is limited to 30K for both languages, covering 97.7\% Chinese words and 99.3\% English words, respectively.

For the WMT En-De translation task, the training set contains 4.5M sentence pairs with 107M English words and 113M German words.
We use the newstest13 and newstest14 as the validation set and test set, respectively.
The joint BPE model is set to 32K merge operations.
\subsection{Setup}
We implement all of the methods based on Transformer \cite{DBLP:journals/corr/VaswaniSPUJGKP17} using the \emph{base} setting with the open-source toolkit \emph{thumt}\footnote{\url{https://github.com/thumt/THUMT}} \cite{Zhang:2017vy}. 
There are six encoder and decoder layers in our models, while each layer employs eight parallel attention heads.
The dimension of the word embedding and the high-level representation $d_{\mathrm{model}}$ is 512, while that of the inner-FFN layer $d_{\mathrm{ff}}$ is 2048.
\begin{table}[t]
\small
    \centering
    \scalebox{1.0}{
    \begin{tabularx}{0.47\textwidth}{@{\extracolsep{\fill}}l|l|lr|l}
        \hline
            & Model &Emb.& Red. &BLEU\\ \hline \hline 
              \multirow{2}{*}{Ar$\Rightarrow$ En}  &Vanilla &23.6M &0\% &28.36\\ 
              &\emph{Shared-private} &11.8M &50\% &29.71$^\uparrow$\\ \hline
              
              \multirow{2}{*}{Ja$\Rightarrow$ En}  &Vanilla &25.6M &0\% &10.94\\ 
              &\emph{Shared-private} &13.3M &48.0\% &12.35$^\uparrow$\\ \hline
              \multirow{2}{*}{Ko$\Rightarrow$ En}  &Vanilla &25.1M &0\% &16.48\\ 
              &\emph{Shared-private} &13.2M &47.4\% &17.84$^\uparrow$\\ \hline
              \multirow{2}{*}{Zh$\Rightarrow$ En}  &Vanilla &27.4M &0\% &19.36\\ 
              &\emph{Shared-private} &13.8M &49.6\% &21.00$^\uparrow$\\ \hline
                                                        
    \end{tabularx}}
    \caption{Results on the IWSLT \{Ar, Ja, Ko, Zh\}-to-En translation tasks. These distant language pairs belonging to 5 different language families and written in 5 different alphabets.``$\uparrow$'' indicates the result is significantly better than that of the vanilla Transformer ($p < 0.01$).}
    \label{tab:iwsltresults}    
\end{table}
The Adam \cite{kingma2014adam} optimizer is used to update the model parameters with hyper-parameters $\beta_{1}$= 0.9, $\beta_{2}$ = 0.98, $\varepsilon$ = $10^{-8}$ and a warm-up strategy with $warmup\_steps = 4000$ is adapted to the variable learning rate \cite{DBLP:journals/corr/VaswaniSPUJGKP17}. 
The dropout used in the residual connection, attention mechanism,  and feed-forward layer is set to 0.1.
We employ uniform label smoothing with 0.1 uncertainty.

During the training, each training batch contains nearly 25K source and target tokens.
We evaluate the models every 2000 batches via the tokenized BLEU \cite{papineni2002bleu} for early stopping.
During the testing, we use the best single model for decoding with a beam of 4. 
The length penalty is tuned on the validation set, which is set to 0.6 for the English-German translation tasks, and 1.0 for others.

We compare our proposed methods with the following related works:
\begin{itemize}
    \item \textbf{Direct bridging} \cite{Kuang:2017vk}: this method minimizes the word embedding loss between the transformations of the target words and their aligned source words by adding an auxiliary objective function.
    \item \textbf{Decoder WT} \cite{Press:2017ug}: this method uses an embedding matrix to represent the target input embedding and target output embedding. 
    \item \textbf{Three-way WT} \cite{Press:2017ug}: this method is an extension of the decoder WT method that the source embedding and the two target embeddings are represented by one embedding matrix. This method cannot be applied to the language pairs with different alphabets, e.g. Zh-En.
\end{itemize}

\begin{table}[t]
\small
    \centering
    \scalebox{1.0}{
    \begin{tabular}{l|rrr|l|l}
        \hline
         Zh-En &$\lambda_\lexical$ &$\lambda_\form$ &$\lambda_\unrelated$   &Emb. &BLEU\\ \hline \hline 
         Vanilla  &-  &- &-  &46.1M &41.37  \\ 
        Decoder WT  &0 &0 &0   &30.7M &41.90  \\ \hline\hline
        \multirow{5}{*}{\emph{Shared-private}}  &0.5 &0.7 &0.9     &21.2M &41.98 \\
        &0.5 &0.5 &0.5    &23.0M &42.26 \\
        &0.9 &0.7 &0 &21.0M &42.27 \\
        &1 &1 &1    &15.3M &42.36 \\
        &0.9 &0.7 &0.5    &18.7M &42.57 \\
        \hline
    \end{tabular}}
    \caption{Performance of models using different sharing coefficients on the validation set of the NIST Chinese-English translation task.} 
    \label{tab:sharecoeff}    
\end{table}

For the proposed model, we use an unsupervised word aligner \emph{fast-align}\footnote{\url{https://github.com/clab/fast_align}} \cite{Dyer:2013wv} to pair source and target words that have similar lexical meaning.
We set the threshold of alignment probability to 0.05, i.e. only those words with an alignment probability over 0.05 can be paired as the words having similar lexical meaning.
The sharing coefficient $\bm{\lambda} = (\lambda_{\lexical},\lambda_{\form},\lambda_{\form})$ is set to (0.9,0.7,0.5), which is tuned on both the NIST Chinese-Enlgish task and the WMT English-German task.


\subsection{Main Results}
Table~\ref{tab:zhenresults} reports the results on the NIST Chinese-English test sets. 
It is observed that the Transformer models significantly outperform SMT and RNNsearch models. 
Therefore, we decide to implement all of our experiments based on Transformer architecture.
The direct bridging model can further improve the translation quality of the Transformer baseline.
The decoder WT model improves the translation quality while reducing the number of parameters for the word representation.
This improved performance happens because there are fewer model parameters, which prevents over-fitting \cite{Press:2017ug}. 
Finally, the performance is further improved by the proposed method while using even fewer parameters than other models.

Similar observations are obtained on the English-German translation task, as shown in Table~\ref{tab:enderesults}.
The improvement of the direct bridging model is reduced with the introduction of sub-word units since the attention distribution of the high-level representations becomes more confused.
Although the two WT methods use fewer parameters, their translation quality degrades.
We believe that sub-word NMT needs the well-trained embeddings to distinguish the homographs of sub-words.
In the proposed method, both the source and target embeddings benefit from the shared features, which leads to better word representations.
Hence, it improves the quality of translation and also reduces the number of parameters.

Table~\ref{tab:iwsltresults} shows the results on the small-scale IWSLT translation tasks. 
We observe that the proposed method stays consistently better than the vanilla model on these distant language pairs.
Although the Three-way WT method has been sufficiently validated on similar translation pairs at low-resource settings \citep{DBLP:journals/corr/SennrichHB16}, it is not applicable to these distant language pairs.
Instead, the proposed method is language-independent, making the WT methods more widely used.

\begin{table}[t]
\small
    \centering
    \scalebox{1.0}{
    \begin{tabular}{l|rrr|l|l}
        \hline
        $A(\cdot|\cdot)$ &Lexical &Form &Unrelated  &Emb. &BLEU \\ \hline \hline 
        0.5 &4,869 &309 &24,822  &22.0M &42.35    \\
        0.1 &15,103 &23 &14,874  &20.0M &42.53  \\ 
        0.05 &21,172 &11  &8,817  &18.7M &42.57 \\  
        \hline
    \end{tabular}}
    \caption{Effects on different alignment thresholds used for pairing the words with similar lexical meaning on the validation set of the NIST Chinese-English translation task.}
    \label{tab:aligneffect}    
\end{table}

\begin{table*}[t]
    \small
    \centering
        \begin{tabular}{p{0.3cm}|p{2cm}|p{12.2cm}}
            \hline 
            \multirow{6}{*}{\emph{1}}& Source                  & mengmai xingzheng zhangguan bazhake biaoshi , dan shi gaishi jiu you shisan \textcolor{red}{\textbf{sangsheng}} .\\
            &Reference               & mumbai municipal commissioner phatak claimed that 13 people were \textcolor{red}{\textbf{killed}} in the city alone .\\\cline{2-3}
            &Vanilla                & bombay chief executive said that there were only 13 deaths in the city alone . \\ 
            &Direct bridging                & bombay 's chief executive , said there were 13 dead in the city alone . \\ 
            &Decoder WT                & chief executive of bombay , said that thirteen people had died in the city alone . \\ 
            &\emph{Shared-private}               & mumbai 's chief executive said 13 people were \textcolor{red}{\textbf{killed}} in the city alone . \\ \hline\hline
            \multirow{6}{*}{\emph{2}} &Source                 & suoyi wo \textcolor{red}{\textbf{ye}} you liyou qu xiangxin ta de rensheng \textcolor{blue}{\textbf{ye}} hen jingcai .\\
            &Reference             & thus , i \textcolor{red}{\textbf{also}} have reason to believe that her life is \textcolor{blue}{\textbf{also}} very wonderful . \\\cline{2-3}
            &Vanilla                & so i have reason to believe her life is \textcolor{blue}{\textbf{also}} very fantastic .\\  
            &Direct bridging                & so i had reason to believe her life was \textcolor{blue}{\textbf{also}} brilliant . \\ 
            &Decoder WT                & so , i have reasons to believe that she has a wonderful life . \\ 
            &\emph{Shared-private}               & so i \textcolor{red}{\textbf{also}} have reason to believe that her life is \textcolor{blue}{\textbf{also}} wonderful . \\ 
             \hline                                                            
        \end{tabular}
    \caption{Translation examples on MT08 test set. The first and second examples show the accuracy and adequacy of the proposed method, respectively. The \textbf{bold} words in each example are paired and will be discussed in the text.}
    \label{tab:translation}
\end{table*}

\subsection{Effect on Sharing Coefficients}
The coefficient $\bm{\lambda}=(\lambda_{\lexical},\lambda_{\form},\lambda_{\unrelated})$ controls the proportion of the shared features.
As shown in Table~\ref{tab:sharecoeff}, the decoder WT model can be seen as a kind of shared-private method where \emph{zero} features are shared between the source and target word embeddings.
For the proposed method, $\bm{\lambda}=(0.5,0.5,0.5)$ and $\bm{\lambda}=(1,1,1)$ are, respectively, used for sharing half and all features between the embeddings of all categories of words.
This allows the model to significantly reduce the number of parameters and also improve the translation quality.
For comparison purpose, we also consider sharing a large part of the features among the unrelated words by setting $s_3$ to $0.9$, i.e. $\bm{\lambda}=(0.5,0.7,0.9)$.
We argue that it is hard for the model to learn an appropriate bilingual vector space in such a sharing setting.

Finally, we propose to share more features between the more similar words by using $s_1=0.9$ and reduce the weight on the unrelated words, which is $\bm{\lambda}=(0.9,0.7,0.5)$.
This strikes the right balance between the translation quality and the number of model parameters.
To investigate whether to share the features between unrelated words or not, we further conduct an experiment with the setting $\bm{\lambda}=(0.9,0.7,0)$.
The result confirms our assumption that a small number of shared features between unrelated words with similar word frequency achieve better model performance.



\begin{figure}[t]%
\centering
\subfigure[Vanilla]{%
\label{fig:ldrfirst}%
\includegraphics[height=1.5in]{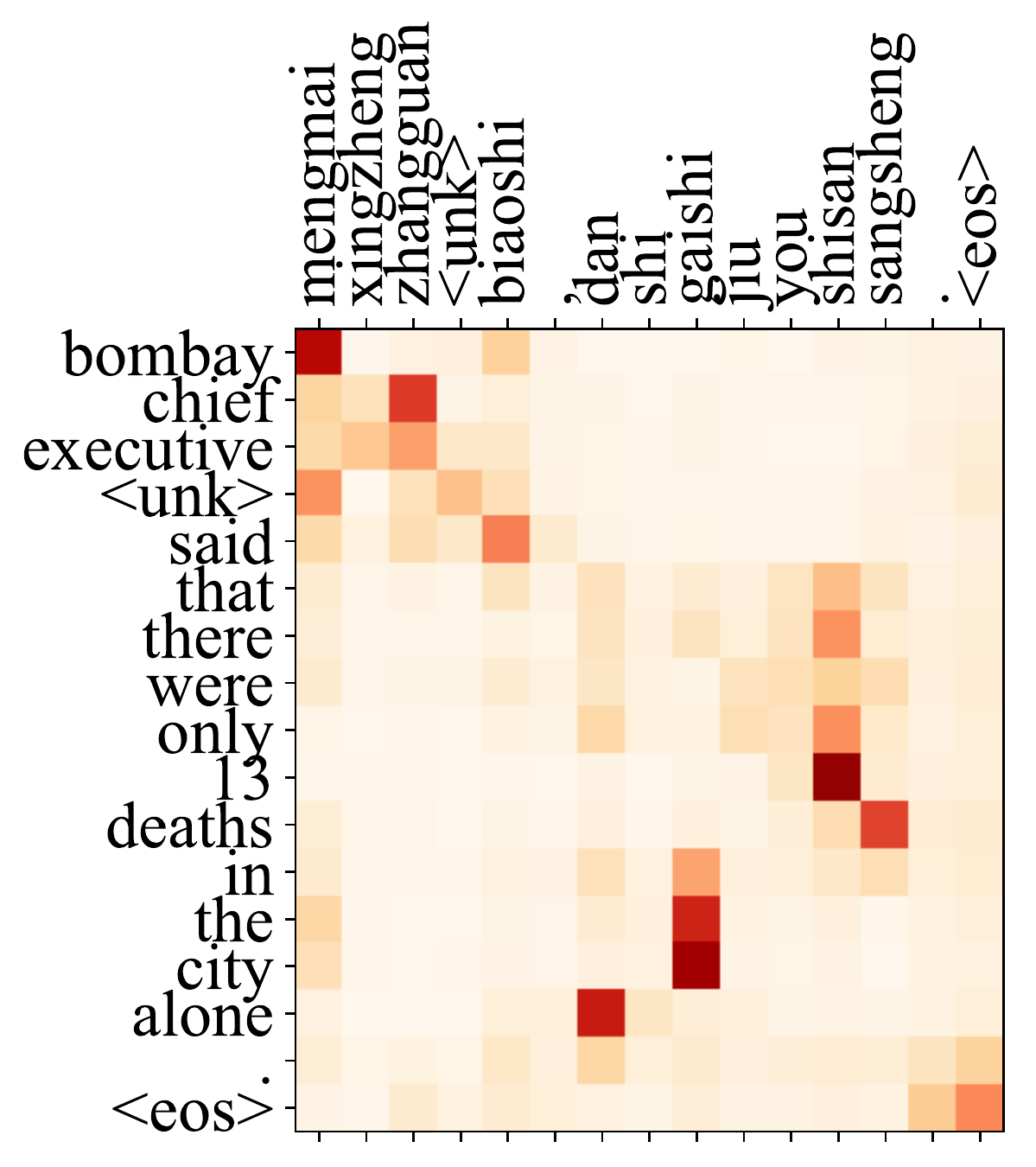}}%
\hfill
\subfigure[Shared-private]{%
\label{fig:ldrthird}%
\includegraphics[height=1.5in]{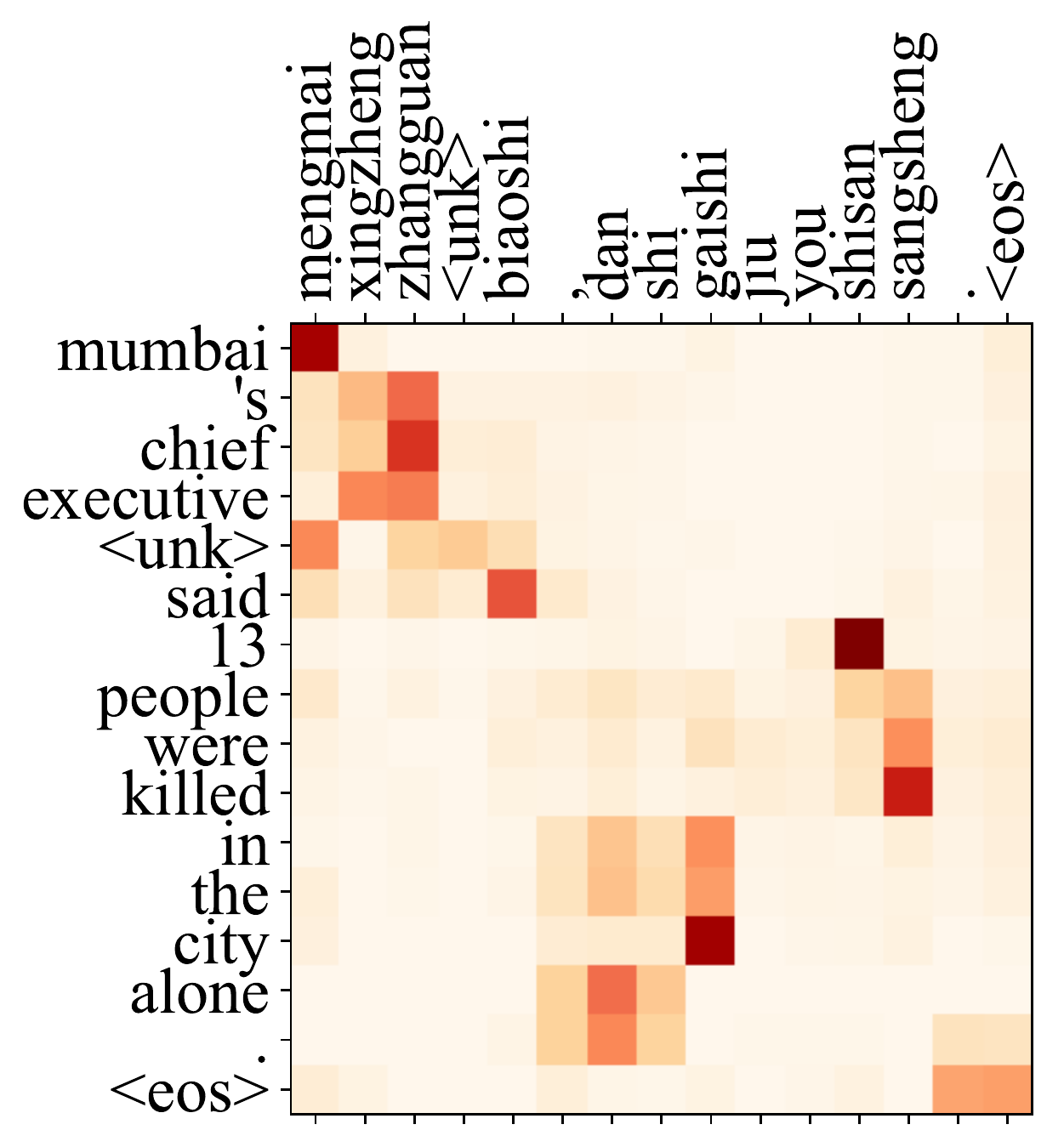}}%
\caption{Long-distance reordering illustrated by the attention maps. The attention weights learned by the proposed shared-private model is more concentrated than that of the vanilla model.}
\label{fig:ldr}%
\end{figure}

\subsection{Effect on Alignment Quality}
Table~\ref{tab:aligneffect} shows the performance of different word alignment thresholds.
In the first row, we only pair the words whose alignment probability $A(y|x)$ is above the threshold of 0.5 (see Equation~\ref{equ:align} for more details).
Under this circumstance, 4,869 words are categorized as parallel words that have similar lexical meaning.
Based on these observations, we find that the alignment quality is not a key factor affecting the model performance.
In contrast, pairing as many as similar words possible helps the model to better learn the bilingual vector space, which improves the translation performance.
The following qualitative analyses support these observations either.

\begin{figure}[t]%
\centering
\subfigure[Vanilla]{%
\label{fig:wofirst}%
\includegraphics[height=1.4in]{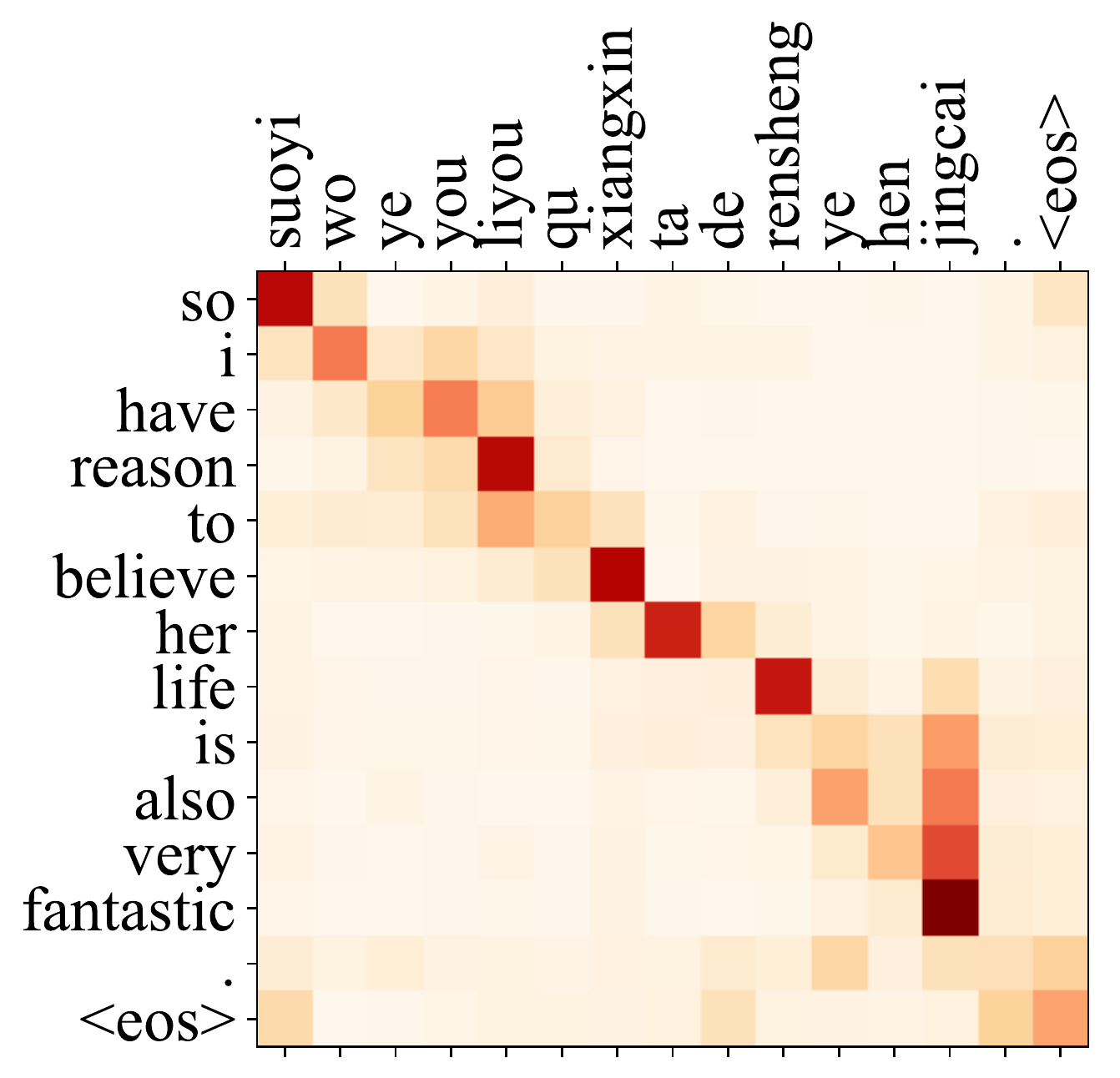}}%
\hfill
\subfigure[Shared-private]{%
\label{fig:wothird}%
\includegraphics[height=1.4in]{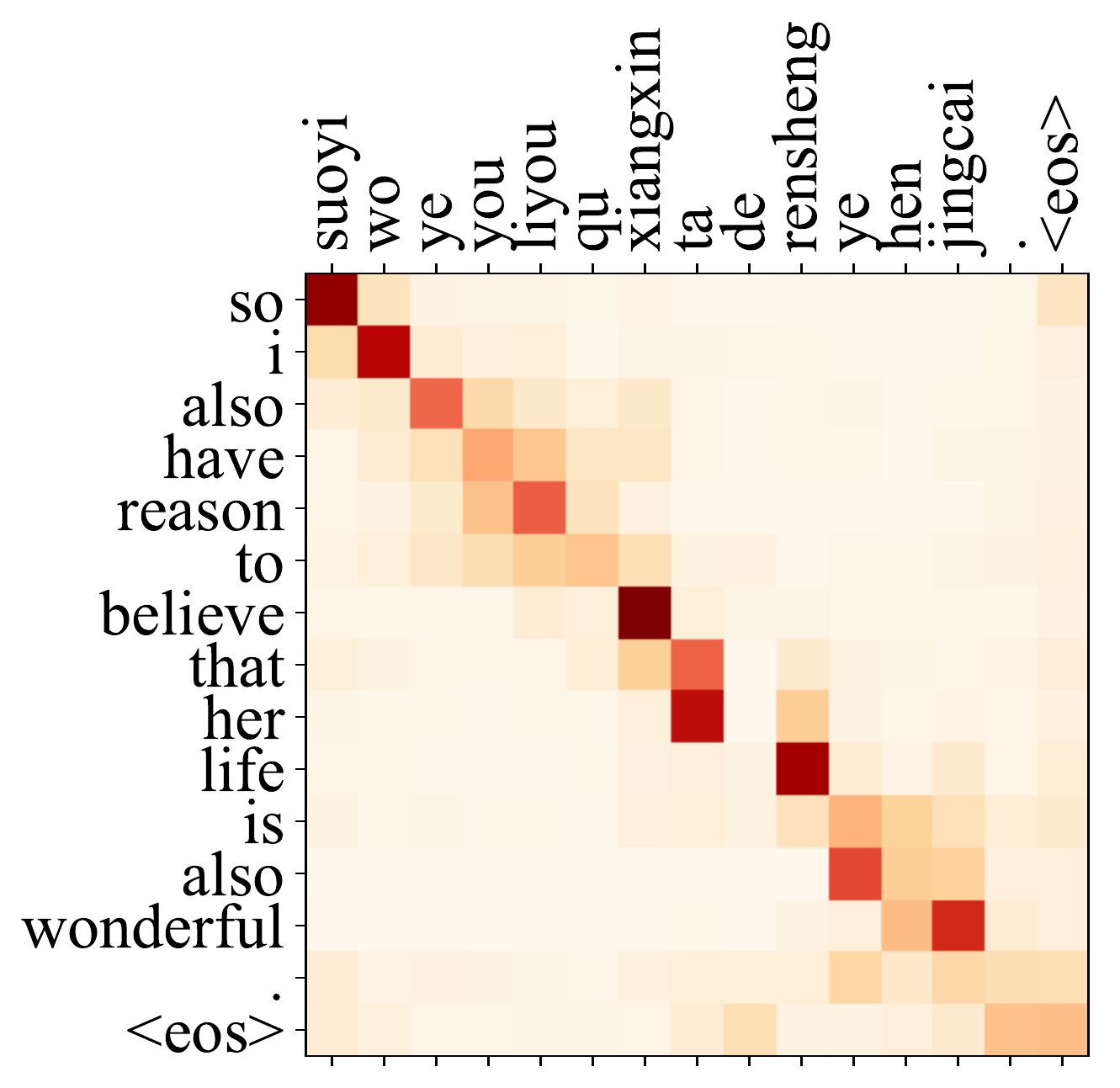}}%
\caption{Word omission problem illustrated by the attention maps. In the vanilla model, the third source word ``ye'' is not translated, while our shared-private model adequately translates it to give a better translation result.}
\label{fig:wo}%
\end{figure}

\begin{figure*}[t]
    \centering
    \subfigure[Vanilla]{
    \scalebox{0.52}{\begin{tikzpicture}
    \begin{axis}
    \Large
        \addplot+[
                visualization depends on={value \thisrow{nodes}\as\myvalue},
                scatter/classes={
                a={mark=text,text mark=\emph{\myvalue},blue},
                b={mark=text,text mark=\myvalue,red}
                },
                scatter,draw=none,
                scatter src=explicit symbolic]
         table[x=x,y=y,meta=label]
            {mydata.dat};
    \end{axis}
\end{tikzpicture}

}
    \label{fig:vanillaWB}}
    \hfill
    \subfigure[Shared-private (global)]{
    \scalebox{0.52}{
\begin{tikzpicture}
	\Large
    \begin{axis}[ytick={0.1,0.2,0.3,0.4}]
        \addplot+[
                visualization depends on={value \thisrow{nodes}\as\myvalue},
                scatter/classes={
                a={mark=text,text mark=\emph{\myvalue},blue},
                b={mark=text,text mark=\myvalue,red}
                },
                scatter,draw=none,
                scatter src=explicit symbolic]     
         table[x=x,y=y,meta=label]{ourdata.dat};
    \end{axis}
\end{tikzpicture}

}
    \label{fig:ourWB}}
    \hfill
    \subfigure[Shared-private (local)]{
    \scalebox{0.52}{
\begin{tikzpicture}
	\Large
    \begin{axis}[ytick={0.25,0.3,0.35},xtick={-0.2,-0.1,0}]
        \addplot+[
                visualization depends on={value \thisrow{nodes}\as\myvalue},
                scatter/classes={
                a={mark=text,text mark=\emph{\myvalue},blue},
                b={mark=text,text mark=\myvalue,red}
                },
                scatter,draw=none,
                scatter src=explicit symbolic]
         table[x=x,y=y,meta=label]
            {ourdatapart.dat};
    \end{axis}
\end{tikzpicture}

}
    \label{fig:ourWBsmall}}
    \caption{Visualization of the 2-dimensional PCA projection of the bilingual word embeddings of the two models. The \emph{\textcolor{blue}{blue}} words represent the Chinese embeddings while the \textcolor{red}{red} words represent the English embeddings. In (a), only the similar monolingual words are clustered together. While in (b) and (c), both the monolingual and bilingual words which have similar meanings are gathered together.}
    \label{fig:WBall}
\end{figure*}
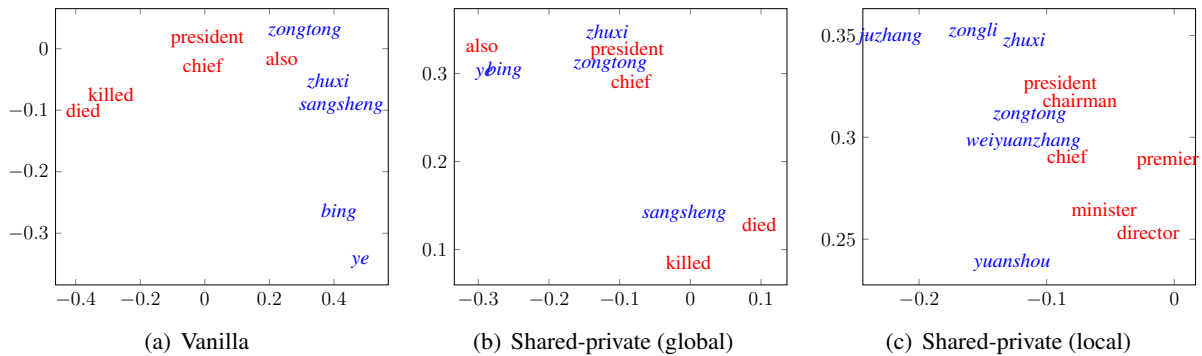

\subsection{Analysis of the Translation Results}
Table~\ref{tab:translation} shows two translation examples of the NIST Chinese-English translation task.
To better understand the translations produced by these two models, we use layer-wise relevance propagation (LRP) \cite{Ding:2017ep} to produce the attention maps of the selected translations, as shown in Figure~\ref{fig:ldr} and~\ref{fig:wo}.

In the first example, the Chinese word ``sangsheng'' is a low-frequency word and its ground truth is ``killed''.
It is observed the inadequate representation of ``sangsheng'' leads to a decline in the translation quality of the vanilla, direct bridging, and decoder WT methods.
In our proposed method, a part of the embedding of ``sangsheng'' is shared with that of ``killed''.
These improved source representations help the model to generate better translations.
Furthermore, as shown in Figure~\ref{fig:ldr}, we observe that the proposed method has better long-distance reordering ability than the vanilla.
We attribute this improvement to the shared features, which provide an alignment guidance for the attention mechanism.

The second example implies that our proposed model is able to improve the adequacy of translation, as illustrated in Figure~\ref{fig:wo}.
The Chinese word ``ye'' (also) appears twice in the source sentence, while only the proposed method can adequately translate both of them to the target word ``also''.
This once again proves that the shared embeddings between the pair words,``ye'' and ``also'' provide the attention model with a strong interaction between the words, leading to a more concentrated attention distribution and effectively alleviating the word omission problem.

\subsection{Analysis of the Learned Embeddings}

The proposed method has a limitation in that each word can only be paired with one corresponding word. 
However, \emph{synonym} is a quite common phenomenon in natural language processing tasks.
Qualitatively, we use principal component analysis (PCA) to visualize the learned embeddings of the vanilla model and the proposed method, as shown in Figure~\ref{fig:WBall}.
In the vanilla model, as shown in Figure~\ref{fig:vanillaWB}, only the similar monolingual embeddings are clustered, such as the English words ``died'' and ``killed'', and the Chinese words ``zhuxi'' (president) and ``zongtong'' (president).
However, in the proposed method, no matter whether the similar source and target words are paired or not, they tend to cluster together; as shown in Figure~\ref{fig:ourWB} and ~\ref{fig:ourWBsmall}.
In other words, the proposed method is able to handle the challenge of synonym.
For example, both the Chinese words ``ye'' (paired with ``also'') and ``bing'' can be correctly translated to ``also'' and these three words tend to gather together in the vector space.
This is similar to the Chinese word ``sangsheng'' (paired with ``killed'') and the English words ``died'' and ``killed''.
Figure~\ref{fig:ourWBsmall} shows that the representations of the Chinese and English words which relate to ``president'' are very close.

\section{Related Work}
Many previous works focus on improving the word representations of NMT by capturing the fine-grained (character) or coarse-grained (sub-word) \emph{monolingual} characteristics, such as character-based NMT \cite{DBLP:journals/corr/Costa-JussaF16,ling2015character,DBLP:journals/corr/ChoMGBSB14,chen2015strategies}, sub-word NMT \cite{DBLP:journals/corr/SennrichHB15,johnson2017google,Ataman:2018wl}, and hybrid NMT \cite{luong2016achieving}.~They effectively consider and utilize the morphological information to enhance the word representations.~Our work aims to enhance word representations through the \emph{bilingual} features that are cooperatively learned by the source and target words.

Recently, \myshortcite{Gu:2018vd} propose to use the pre-trained target (English) embeddings as a universal representation to improve the representation learning of the source (low-resource) languages. 
In our work, both the source and target embeddings can make use of the common representation unit, i.e. the source and target embedding help each other to learn a better representation.

The previously proposed methods have shown the effectiveness of integrating prior word alignments into the attention mechanism \cite{Mi:2016wd,Liu:2016ta,Cheng:2016uy,Feng:vt}, leading to more accurate and adequate translation results with the assistance of prior guidance.
We provide an alternative that integrates the prior alignments through the sharing of features, which can also leads to a reduction of model parameters.



\myshortcite{Kuang:2017vk} propose to shorten the path length between the related source and target embeddings to enhance the embedding layer.
We believe that the shared features can be seem as the \emph{\textbf{zero}} distance between the paired word embeddings.
Our proposed method also uses several ideas from the three-way WT method~\citep{Press:2017ug}.
Both of these methods are easy to implement and transparent to different NMT architectures.
The main differences are: 1) we share a part of features instead of all features; 2) the words of different relationship categories are allowed to share with differently sized features; and (3) it is adaptable to any language pairs, making the WT methods more widely used.

\section{Conclusion}
In this work, we propose a novel sharing technique to improve the learning of word embeddings for NMT. 
Each word embedding is composed of shared and private features. 
The shared features act as a prior alignment guidance for the attention model to improve the quality of attention. 
Meanwhile, the private features enable the words to better capture the monolingual characteristics, result in an improvement of the overall translation quality. 
According to the degree of relevance between a parallel word pair, the word pairs are categorized into three different groups and the number of shared features is different. 
Our experimental results show that the proposed method outperforms the strong Transformer baselines while using fewer model parameters.


\section*{Acknowledgements}
This work is supported in part by the National Natural Science Foundation of China (Nos. 61672555, 61876035, 61732005), the Joint Project of Macao Science and Technology Development Fund and National Natural Science Foundation of China (No. 045/2017/AFJ),  the Multi-Year Research Grant from the University of Macau (No. MYRG2017-00087-FST).
Yang Liu is supported by the National Key R\&D Program of
China (No. 2017YFB0202204), National Natural Science Foundation of China (No. 61761166008, No. 61432013), Beijing Advanced Innovation Center for Language Resources (No. TYR17002).

\bibliography{acl2019}
\bibliographystyle{acl_natbib}

\clearpage

\end{document}